\PassOptionsToPackage{dvipsnames}{xcolor}
\documentclass[11pt,letterpaper]{style}

\usepackage{booktabs}
\usepackage{longtable}
 \usepackage{hyperref}
\usepackage{url}
\usepackage{multirow}
\usepackage{listings}
\usepackage[most]{tcolorbox}
\usepackage{booktabs}
\usepackage{microtype}
\usepackage{authblk} 
\usepackage{amsmath}
\usepackage{amssymb}
\usepackage{colortbl}   
\usepackage{xcolor}    
\usepackage{listings}
\usepackage{algorithm2e}
\usepackage{natbib}

\SetKwInOut{KwData}{Data}
\SetKwInOut{KwInput}{Input}
\SetKwInOut{KwOutput}{Output}

\title{\centering From Attribution to Abstention: \\Training-Free Attention-Based Auditing for Clinical Summarization}
\runningtitle{From Attribution to Abstention: Training-Free Attention-Based Auditing for Clinical Summarization}

\author{%
    \textbf{Qianqi Yan}\textsuperscript{1} \quad
    \textbf{Huy Nguyen}\textsuperscript{2} \quad
    \textbf{Sumana Srivatsa}\textsuperscript{2} \quad
    \textbf{Hari Bandi}\textsuperscript{2} \quad
    \textbf{Xin Eric Wang}\textsuperscript{1} \quad
    \textbf{Krishnaram Kenthapadi}\textsuperscript{2}
    \\
    {\normalfont\small
    \textsuperscript{1} University of California, Santa Barbara \quad
    \textsuperscript{2} Oracle Health AI
    }
}

\begin{document}

\begin{abstract}
Deploying multimodal large language models (MLLMs) for clinical summarization demands not only fluent generation but also transparency about where each statement originates---and a mechanism to flag when statements lack evidential support.
We present \textsc{ClinTrace}, a training-free framework that extracts two clinically useful signals from the decoder attention weights that every transformer-based MLLM already produces during generation:
(i)~fine-grained \emph{source attributions} linking each output sentence to supporting text spans or images, and
(ii)~per-sentence \emph{groundedness scores} that identify poorly supported claims as candidate hallucinations.
Both signals are derived from the same attention tensors in a single pass, requiring no retraining, no auxiliary models, and no additional inference cost.
We evaluate on two clinical summarization tasks---doctor--patient dialogue summarization (\textsc{CliConSummation}) and radiology report summarization (\textsc{MIMIC-CXR})---using a general-purpose MLLM (\texttt{Qwen3-8B}) and a medical-finetuned model (\texttt{HuatuoGPT-Vision-7B}).
For source attribution, \textsc{ClinTrace} achieves over 92\% text F1 on radiology and 88\% on dialogue summarization, substantially outperforming embedding-based and self-attribution baselines.
For hallucination detection, groundedness scores achieve 0.77 AUROC with the medical-finetuned model---competitive with embedding-based confidence at zero additional cost---and enable an abstention mechanism that improves faithfulness from 61.7\% to 72.6\% by withholding the least-grounded 20\% of output for clinician review.
Notably, medical finetuning substantially improves the reliability of attention-based hallucination detection, suggesting that domain adaptation produces more semantically structured attention patterns amenable to self-auditing.
\end{abstract}

\maketitle

\begin{figure}[!ht]
\setlength{\abovecaptionskip}{0.5cm}
    \centering
    \includegraphics[width=\columnwidth]{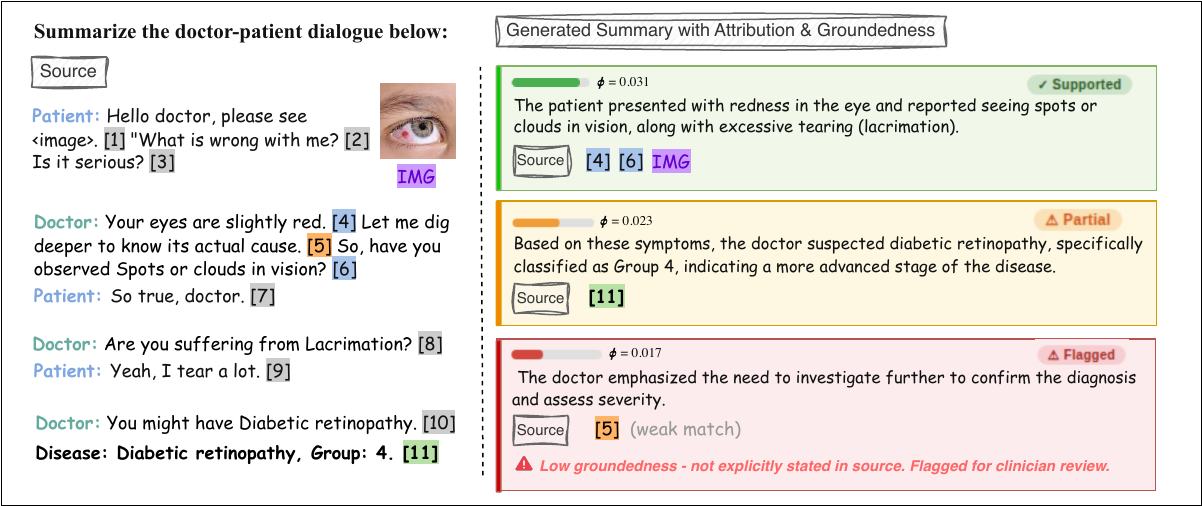}
    \caption{\textbf{Overview of \textsc{ClinTrace}.}
    Given a clinical input---here, a doctor--patient dialogue with an embedded image of the patient's eye---our framework extracts decoder attention weights during standard generation to produce:
    \textbf{(1)}~source attributions linking each generated sentence to supporting text spans and the image (colored citation tags);
    \textbf{(2)}~per-sentence groundedness scores $\phi$ reflecting how strongly each sentence is grounded in the source (horizontal bars); and
    \textbf{(3)}~an abstention decision that flags the least-grounded sentence for clinician review.
    Sentence~3 (``The doctor emphasized the need to investigate further...'') receives the lowest groundedness score ($\phi{=}0.017$) and is correctly flagged: the source dialogue contains no such statement---the model fabricated it.
    All three signals are derived from the same attention tensors at zero additional inference cost.}
    \label{fig:teaser}
\end{figure}

\section{Introduction}
\label{sec:intro}

Multimodal large language models (MLLMs) are increasingly applied to clinical 
documentation tasks---distilling lengthy doctor--patient dialogues into concise notes and summarizing radiology findings into 
impressions~\citep{van2024adapted, doshi2024quantitative, thawakar2024xraygpt, alsaad2024survey}.
These systems promise to reduce documentation burden and accelerate information retrieval for clinicians.
Recent instruction-tuned models produce highly fluent outputs, often rivaling supervised baselines on automatic metrics~\citep{gpt-5.2, anthropic2025claude, gemini3, bai2025qwen3}.

However, fluency alone is insufficient for clinical deployment.
MLLMs remain prone to \emph{hallucinations}: claims that are weakly supported or entirely unsupported by the underlying evidence~\citep{gu2026medvh, yan2024medhvl, 10.1145/3571730}.
In high-stakes domains such as radiology and patient documentation, an unsupported statement about a lesion or a misattributed patient complaint can lead to delayed diagnosis, unnecessary tests, or erosion of clinician trust~\citep{alsaad2024survey}.
For real-world adoption, practitioners require two complementary capabilities: \emph{traceability}---a clear indication of where each statement originated---and \emph{reliability monitoring}---a mechanism to flag when a generated statement may not be trustworthy.
Together, these form a \emph{clinical audit layer} that enables clinicians to efficiently verify model outputs before acting on them.

Existing approaches to transparency and hallucination mitigation in MLLMs fall short of what clinical deployment demands.
Post-hoc attribution methods---including gradient-based saliency~\citep{barkan2021gradsam, selvaraju2020grad}, attention rollout~\citep{attcat, abnar-zuidema-2020-quantifying}, and relevance propagation~\citep{song-etal-2024-better, bach2015pixel}---were developed for classification and encoder-based architectures, where explanations are computed
with respect to a fixed objective such as a class logit.
They do not naturally extend to open-ended generation, where attribution must operate over a growing causal sequence with no externally specified target.
Training-based hallucination mitigation methods such as RLHF~\citep{ouyang2022training} and contrastive decoding~\citep{li2023contrastive} can reduce hallucination rates but require
expensive retraining, do not provide per-sentence provenance, and offer no mechanism for a clinician to verify \emph{why} a particular statement was generated. 
Retrieval-augmented approaches~\citep{shuster2021retrieval} ground generation in external knowledge but do not trace outputs back to the input document itself. 
Meanwhile, medical-specific hallucination benchmarks~\citep{gu2026medvh, yan2024medhvl} have documented the severity of the problem but focus on \emph{evaluation} rather than providing actionable signals during deployment. 
There remains a gap between what MLLMs produce and the kind of accountable, per-sentence auditable outputs that clinical workflows demand---outputs where each claim is linked to its source and flagged when evidential support is weak.

In this work, we investigate a simpler question: \textbf{do the attention weights that every transformer-based MLLM already computes during generation contain enough signal to serve as a clinical audit layer?}
We find that the answer is yes---with caveats that are themselves clinically informative.

We present \textsc{ClinTrace}, a training-free framework that extracts two signals from decoder attention tensors during standard autoregressive generation:
(1)~\emph{source attributions} that map each generated sentence to the input spans and images it draws from, and
(2)~\emph{groundedness scores} that quantify how confidently each sentence can be traced back to source evidence, serving as a hallucination risk indicator.
Both signals are derived from the same attention weights in a single pass, with no retraining, no auxiliary models, and no additional inference cost beyond enabling attention output.

We evaluate \textsc{ClinTrace} on two clinical summarization tasks---doctor--patient dialogue summarization (\textsc{CliConSummation}~\citep{tiwari2023experience}), including both text-only and multimodal encounters, and radiology report summarization (\textsc{MIMIC-CXR}~\citep{johnson2019mimic})---using two MLLM backbones: \texttt{Qwen3-8B}~\cite{bai2025qwen3}, a general-purpose model, and \texttt{HuatuoGPT-Vision-7B}~\citep{chen2024towards}, a medical-domain model fine-tuned on PubMedVision data using the Qwen2.5-VL~\citep{bai2025qwen2} architecture.
This pairing enables a controlled comparison of how general-purpose and medical-finetuned models differ in the clinical utility of their attention patterns.

Our findings include:

\begin{enumerate}
    \item \textbf{Attention-based attribution is highly effective.}
    With principled aggregation (pooling across layers and heads, sentence-level majority voting), decoder attentions yield source citations that substantially outperform embedding-based and self-attribution baselines---achieving over 90\% text F1 on both radiology and dialogue summarization (Table~\ref{tab:attribution}).

    \item \textbf{Groundedness scores detect hallucinations---but model choice matters.}
    Per-sentence groundedness scores derived from the same attention signal achieve 0.77 AUROC for hallucination detection with the medical-finetuned model (HuatuoGPT), competitive with embedding-based confidence at zero additional cost.
    For the general-purpose model (Qwen3), the signal is weaker (0.59 AUROC), suggesting that medical finetuning produces more semantically structured attention patterns that are more amenable to self-auditing.

    \item \textbf{Abstention improves reliability for medical-finetuned models.}
    Using groundedness scores to withhold the least-grounded 20\% of output improves faithfulness from 61.7\% to 72.6\% for HuatuoGPT-Vision-7B.
    This demonstrates a practical deployment strategy: present high-confidence output normally while flagging low-confidence sentences for clinician review.

    \item \textbf{Attribution, groundedness, and faithfulness are connected.}
    Correctly attributed sentences have systematically higher groundedness scores and higher faithfulness rates than incorrectly attributed ones (Figure~\ref{fig:connection}), confirming that these are three views of the same underlying signal.
\end{enumerate}

\subsection*{Generalizable Insights about Machine Learning in the Context of Healthcare}

This work provides several insights that extend beyond the specific tasks and datasets studied:

\begin{itemize}
    \item \textbf{Decoder attentions as a free clinical resource.}
    Every transformer-based MLLM already produces attention weights during generation, yet these are routinely discarded.
    We show that with principled aggregation, these signals yield both reliable source citations and hallucination detectors---without any retraining, additional inference cost, or auxiliary models.
    This suggests a general recipe applicable to any clinical NLP pipeline built on transformer decoders.

    \item \textbf{Medical finetuning improves attention interpretability.}
    A key finding is that medical-finetuned models produce attention patterns that are substantially more useful for self-auditing than those of general-purpose models.
    This has practical implications for clinical deployment: when choosing between a general-purpose and a domain-adapted MLLM, the latter offers not only better task performance but also more reliable built-in safety signals.

    \item \textbf{Abstention over silent failure.}
    In clinical AI, a system that \emph{knows when it doesn't know} is more valuable than one that always produces an answer.
    Our groundedness-based abstention mechanism exemplifies a design principle for clinical ML: rather than attempting to silently correct errors, surface uncertainty to the clinician and let them decide.
\end{itemize}
\section{Related Work}
\label{sec:related}

\paragraph{Attribution Methods for Generative Models.}
Post-hoc attribution for neural networks has been extensively studied, with most methods developed for classification and encoder-based architectures where explanations target a fixed objective such as a class logit~\citep{song-etal-2024-better, attcat, abnar-zuidema-2020-quantifying, selvaraju2020grad, voita-etal-2019-analyzing, bach2015pixel}.
These approaches produce token-level importance scores via attention analysis, gradient-based saliency, or relevance propagation.
While effective in constrained settings, they do not directly address attribution in open-ended generation, where there is no fixed target and the causal graph grows at each decoding step.

For generative models, linking outputs to provenance has gained traction through several complementary directions.
Faithfulness evaluation frameworks such as SummaC~\citep{laban-etal-2022-summac} use NLI-style consistency checks between summaries and sources, while SelfCheckGPT~\citep{manakul-etal-2023-selfcheckgpt} estimates hallucination risk via self-consistency across multiple samples.
More directly, Traceable Text~\citep{kambhamettu2024traceable} introduces phrase-level provenance links attaching source spans to generated spans, and \citet{suhara-alikaniotis-2024-source} formulates sentence-level source identification for abstractive summarization.
Concurrent work on generation-time attribution, OmniTrace~\citep{yan2026omnitraceunifiedframeworkgenerationtime}, introduces a unified framework that converts token-level attribution signals into span-level source explanations across text, image, audio, and video modalities in decoder-only MLLMs. Our work instantiates a similar attention-based attribution pipeline but focuses specifically on the clinical setting, where we extend the framework to serve a dual purpose: source citation and hallucination detection via groundedness scoring.
Our work differs from these approaches in two ways: we operate \emph{during} generation rather than post-hoc, and we show that the same attention signal used for attribution also yields a hallucination detection signal---connecting provenance tracking and reliability monitoring within a single framework.

\paragraph{Clinical Summarization and Transparency.}
Clinical summarization raises uniquely high-stakes requirements for faithfulness.
Surveys of radiology report generation~\citep{sloan2024automated} emphasize persistent gaps in factual grounding and call for methods that surface supporting evidence.
Task-driven work in radiology, such as pragmatic report generation conditioned on indications and prior studies~\citep{nguyen2023pragmatic}, demonstrates that integrating structured context improves clinical relevance, but interpretability remains post-hoc or coarse-grained.
Structured evaluation resources such as RadGraph~\citep{jain2021radgraph} enable entity-level assessment of factual correctness, and DocLens~\citep{zhu2025doclens} includes attribution as an evaluation dimension.
However, none of these provide \emph{generation-time} attribution---they evaluate outputs after the fact rather than producing citations alongside the generated text.
Recent work has shown that adapted LLMs can match or exceed medical experts on clinical summarization~\citep{van2024adapted}, making the transparency gap increasingly urgent: as models become more capable, the need for clinicians to verify \emph{why} a model produced a particular statement grows correspondingly.

\paragraph{Hallucination Detection in Medical MLLMs.}
Hallucination in multimodal LLMs has received growing attention, with benchmarks documenting the scope of the problem in clinical settings.
MedVH~\citep{gu2026medvh} introduces a comprehensive benchmark across five tasks, revealing that even domain-finetuned models produce hallucinated responses that fail to reflect input medical images.
Med-HVL~\citep{yan2024medhvl} defines object hallucination and domain knowledge hallucination metrics for medical LVLMs and proposes an automated evaluation framework.
\citet{chen2024detecting} propose methods for detecting and evaluating medical hallucinations in large vision-language models, while RadFlag~\citep{zhang2024radflag} offers a black-box detection method specifically for medical VLMs.
On the mitigation side, training-based approaches such as RLHF~\citep{ouyang2022training} and contrastive decoding~\citep{li2023contrastive} can reduce hallucination rates, and retrieval-augmented generation~\citep{shuster2021retrieval} grounds outputs in external knowledge.
However, these methods either require expensive retraining, do not provide per-sentence provenance, or ground in external rather than input-document evidence.
Our approach is complementary: rather than modifying training or decoding, we use attention patterns already present during standard generation as a hallucination signal, requiring no additional computation.
A key finding of our work is that the effectiveness of this signal depends on whether the model has been finetuned on medical data---an observation that connects hallucination detection to model selection decisions in clinical deployment.
\section{ClinTrace}
\label{sec:method}

Our framework extracts two clinically useful signals from the decoder attention weights that every transformer-based MLLM already produces during generation:
(1)~\emph{source attribution}---which input spans support each generated sentence, and
(2)~\emph{groundedness scoring}---how confidently each generated sentence can be traced back to the source, serving as a proxy for hallucination risk.
Both signals are derived from the same attention tensors in a single pass, requiring no retraining, no auxiliary models, and no additional inference cost beyond enabling attention output.

\subsection{Task Formulation}
\label{sec:formulation}

Given a source document $D$ which may be text-only (a radiology report, a clinical dialogue) or multimodal (dialogue with an embedded image), and a generated output $G$ (a summary, for example), we seek two outputs for each generated sentence $g_j \in G$:

\begin{enumerate}
    \item \textbf{Attribution} $\mathcal{A}(j) \subseteq \{s_1, \dots, s_{|S|}\} \cup \{s_{\text{img}}\}$: the set of source sentences (and optionally the image) that support $g_j$.
    \item \textbf{Groundedness score} $\phi(j) \in \mathbb{R}$: a scalar reflecting how strongly $g_j$ is grounded in the source, where low values indicate candidate hallucinations.
\end{enumerate}

Both are computed from the same intermediate representation: the pooled attention distribution from each generated token to the source tokens.

\subsection{Attention-Based Attribution and Groundedness}
\label{sec:pipeline}

The pipeline operates in three stages, illustrated in Algorithm~\ref{alg:clintrace}.
The first two stages---chunking and attention pooling---produce a per-token attention signal.
The third stage aggregates this signal into sentence-level attributions \emph{and} groundedness scores simultaneously.

\paragraph{Stage 1: Source chunking.}
The source document is segmented into sentence-level evidence units, and each source token is assigned to its enclosing sentence.
For multimodal inputs, image patch tokens form a contiguous block and are treated as a single evidence unit $s_{\text{img}}$.
This sentence-level granularity aligns with how clinicians naturally reason about clinical notes: each source sentence corresponds to a discrete clinical observation, complaint, or finding.

\paragraph{Stage 2: Attention pooling.}
For each generated token $y_t$, we extract its attention weights over all source tokens across every layer and head, then average to obtain a single pooled vector $\bar{\mathbf{a}}_t \in \mathbb{R}^K$:
\[
\bar{\mathbf{a}}_t = \text{mean}_{\ell, h}\bigl(A_t^{(\ell, h)}\bigr),
\]
where $A_t^{(\ell, h)}$ is the attention distribution from token $y_t$ at layer $\ell$, head $h$.
Averaging across all layers and heads serves as a denoising mechanism: individual heads may capture syntactic, positional, or semantic patterns, but their average provides a stable estimate of overall source relevance that generalizes across model architectures~\citep{abnar-zuidema-2020-quantifying}.

\paragraph{Stage 3: Attribution and groundedness.}
From the pooled attention vector $\bar{\mathbf{a}}_t$, we derive two quantities per generated token:

\emph{Source assignment.}
Each token is mapped to its most-attended source sentence by summing pooled attention within each source span and selecting the maximum:
\[
\hat{s}(t) = \arg\max_{S_j \in \mathcal{S}} \sum_{i \in S_j} \bar{a}_t(i).
\]

\emph{Token-level groundedness.}
The confidence of this assignment is the corresponding maximum summed attention:
\[
c_t = \max_{S_j \in \mathcal{S}} \sum_{i \in S_j} \bar{a}_t(i).
\]
A high $c_t$ indicates that token $y_t$ attends strongly to a specific source span; a low $c_t$ indicates diffuse attention with no clear provenance.

At the sentence level, we aggregate across all tokens $T_j$ in generated sentence $g_j$:

\emph{Sentence attribution.}
The set of source sentences that receive consistent support from $g_j$'s tokens forms the attribution:
\[
\mathcal{A}(j) = \bigl\{ s : \text{count}(\hat{s}(t) = s,\; t \in T_j) \geq \tau \cdot |T_j| \bigr\},
\]
where $\tau$ is a threshold controlling the minimum fraction of tokens that must agree on a source for it to be retained as a citation.

\emph{Sentence groundedness.}
The groundedness score aggregates token-level confidences:
\[
\phi(j) = f\bigl(\{c_t : t \in T_j\}\bigr),
\]
where $f$ is an aggregation function.
Intuitively, $\phi(j)$ measures how concentrated the attention of $g_j$'s tokens is on specific source evidence: high concentration suggests the sentence is well-grounded, while diffuse attention suggests the model is generating without clear evidential support.

\paragraph{Image attribution.}
For multimodal inputs, we additionally compute each generated token's total attention to the image patch block:
$w_{\text{img}}(t) = \sum_{i \in \mathcal{I}} \bar{a}_t(i)$,
where $\mathcal{I}$ is the set of image token indices.
At the sentence level, we average these weights and attribute the image to all sentences whose mean image attention exceeds a threshold $\tau_{\text{img}}$.

\begin{algorithm}[t]
\small
\SetAlgoLined
\SetAlgoNlRelativeSize{0}
\DontPrintSemicolon
\caption{ClinTrace: Attention-Based Attribution and Groundedness Scoring}
\label{alg:clintrace}
\vspace{10pt}
\KwData{Source $D$ (text $\pm$ image), model $M$, threshold $\tau$, aggregation function $f$}
\KwOutput{Generated text $\mathbf{y}$, per-chunk attributions $\mathcal{A}$, groundedness scores $\phi$}

\vspace{0.3em}
\tcc{Preprocessing}
$\mathbf{x} \leftarrow \texttt{Tokenize}(D)$\;
$\mathcal{S} \leftarrow \texttt{ChunkSources}(\mathbf{x}, D)$ \tcp*{sentence-level source units + image block}

\vspace{0.3em}
\tcc{Generation with attention capture}
$\mathbf{y},\; \{A_t^{(\ell,h)}\} \leftarrow M.\texttt{generate}(\mathbf{x},\; \texttt{output\_attentions}=\texttt{True})$\;

\vspace{0.3em}
\tcc{Token-level attribution and groundedness}
\ForEach{generated token $y_t$}{
    $\bar{\mathbf{a}}_t \leftarrow \mathrm{mean}_{\ell,h}\!\bigl(A_t^{(\ell,h)}\bigr)$ \tcp*{pool across layers and heads}
    $\hat{s}(t) \leftarrow \arg\max_{S_j \in \mathcal{S}} \sum_{i \in S_j} \bar{a}_t(i)$ \tcp*{source assignment}
    $c_t \leftarrow \max_{S_j \in \mathcal{S}} \sum_{i \in S_j} \bar{a}_t(i)$ \tcp*{token groundedness}
}

\vspace{0.3em}
\tcc{Sentence-level aggregation}
$\mathcal{C} \leftarrow \texttt{ChunkOutput}(\mathbf{y})$ \tcp*{segment generation into sentences}
\ForEach{chunk $C_k \in \mathcal{C}$ with token set $T_k$}{
    $\mathcal{A}(k) \leftarrow \bigl\{ s \in \mathcal{S} : \frac{|\{t \in T_k : \hat{s}(t) = s\}|}{|T_k|} \geq \tau \bigr\}$ \tcp*{attribution}
    $\phi(k) \leftarrow f\!\bigl(\{c_t : t \in T_k\}\bigr)$ \tcp*{groundedness score}
    \If{multimodal input}{
        $w_k \leftarrow \frac{1}{|T_k|} \sum_{t \in T_k} \sum_{i \in \mathcal{I}} \bar{a}_t(i)$ \tcp*{image attention}
        \If{$w_k \geq \tau_{\mathrm{img}}$}{
            $\mathcal{A}(k) \leftarrow \mathcal{A}(k) \cup \{s_{\mathrm{img}}\}$\;
        }
    }
}

\vspace{0.3em}
\Return $\mathbf{y}$, $\{\mathcal{A}(k)\}_{k=1}^{|\mathcal{C}|}$, $\{\phi(k)\}_{k=1}^{|\mathcal{C}|}$\;
\end{algorithm}

The key observation motivating our hallucination detection approach is that \emph{the same attention signal used for attribution also encodes how well-supported each generated sentence is}.
When a model generates a faithful sentence, its tokens tend to attend strongly to specific source spans, producing high groundedness scores and correct attributions.
When a model hallucinates, attention is typically more diffuse---no single source span dominates---resulting in low groundedness and unreliable attribution.

We operationalize this observation through \emph{abstention}: generated sentences with groundedness scores below a threshold $\phi_{\min}$ are flagged for clinician review rather than presented as confident model output.
This design reflects a clinical safety principle: rather than attempting to silently correct errors, the system surfaces uncertainty and defers to the clinician.
The threshold $\phi_{\min}$ controls the trade-off between output coverage (how much of the summary is retained) and faithfulness (what fraction of retained sentences are factually supported).
We evaluate this trade-off empirically in Section~\ref{sec:results_abstention}.
\section{Cohort}
\label{sec:cohort}

We evaluate on two publicly available clinical datasets that span distinct documentation settings: informal multimodal dialogues and formal radiology prose.

\subsection{Cohort Selection}

\paragraph{CliConSummation}~\citep{tiwari2023experience} consists of doctor--patient dialogues drawn from online clinical consultation platforms.
Each dialogue involves a patient describing symptoms and a doctor providing assessment, with some encounters including clinical images shared by the patient (e.g., photographs of skin conditions).
We use the full provided split, comprising 1{,}405 text-only dialogues and 257 multimodal dialogues (each containing a single embedded image), for a total of 1{,}662 samples.
No additional filtering or exclusion criteria were applied beyond the dataset's original curation.

\paragraph{MIMIC-CXR}~\citep{johnson2019mimic} is a large-scale dataset of chest radiographs paired with free-text radiology reports from the Beth Israel Deaconess Medical Center (2011--2016).
We use the standard \textsc{Findings}~$\rightarrow$~\textsc{Impression} summarization task: given the Findings section of a radiology report, the model generates the Impression (a concise summary of key observations).
We filter for reports containing both a non-empty Findings section ($\geq$3 sentences) and a non-empty Impression, using the official MIMIC-CXR test split.
This yields 1{,}603 text-only samples.
Access was obtained under the PhysioNet Credentialed Health Data License.

\subsection{Data Extraction}

For both datasets, source documents are segmented into sentence-level evidence units using rule-based sentence tokenization.
Each source token is assigned to its enclosing sentence to enable token-to-sentence mapping during attribution.
For multimodal dialogues in CliConSummation, the image is retained as a contiguous block of patch tokens in the model's input sequence and treated as a single evidence unit.

Generated summaries are produced by prompting each model with the source document (and image, when present) and a task-specific instruction.
For CliConSummation, the prompt instructs the model to summarize key clinical information including the chief complaint, symptoms, and the doctor's assessment.
For MIMIC-CXR, the prompt instructs the model to generate the Impression section from the Findings.
All generation uses greedy decoding to ensure reproducibility.
Generated outputs are segmented into sentence-level chunks using the same tokenizer for downstream attribution and faithfulness evaluation. Prompt details can be found in Appendix~\ref{app:task_prompt}.
\section{Experiments}
\label{sec:results}

\subsection{Experimental Setup}
\label{sec:setup}

We evaluate our framework along three axes:
(1)~\emph{source attribution}: can attention-based semantic tracing outperform strong baselines?
(2)~\emph{hallucination detection}: do the same attention signals predict faithfulness?
(3)~\emph{clinical utility}: can groundedness-based abstention improve the reliability of retained output?
All experiments span two clinical domains (radiology report summarization on \textsc{MIMIC-CXR}~\citep{johnson2019mimic} and doctor--patient dialogue summarization on \textsc{CliConSummation}~\citep{tiwari2023experience}, totaling 3{,}265 samples) and two model backbones: \texttt{Qwen3-8B}~\citep{bai2025qwen3}, a general-purpose MLLM, and \texttt{HuatuoGPT-Vision-7B}~\citep{chen2024towards}, a medical-domain model fine-tuned on the PubMedVision dataset using the Qwen2.5-VL~\citep{bai2025qwen2} architecture.

\paragraph{Evaluation protocol.}
For each generated summary, we segment both the source document and the generated output into sentence-level chunks.
Attribution quality is measured by comparing the predicted source set for each generated chunk against a reference source set, using two metrics:
\emph{Text~F1}, the macro-averaged F1 between predicted and reference source sentence sets (treating each source sentence as a label in a multi-label classification), and
\emph{Text~EM} (exact match), the fraction of generated chunks whose predicted source set exactly matches the reference.
For multimodal inputs, we additionally report \emph{Image~F1}, measuring whether the image is correctly included in or excluded from the predicted source set.

Reference source sets are constructed by prompting a separate LLM (Claude Opus 4.6~\citep{anthropic2025claude}) to identify, for each generated chunk, which source sentences (and the image, if present) support it.
To reduce annotation bias, the judge model is distinct from both generation models, and prompt templates are fixed across all conditions (details in Appendix~\ref{app:llm_judge_prompt}).
We validated judge reliability on a random subset of 100 chunks through manual inspection, finding agreement rates consistent with prior work using LLM-based evaluation in clinical NLP~\citep{zhu2025doclens, van2024adapted}.

\subsection{Source Attribution}
\label{sec:results_attr}

Table~\ref{tab:attribution} reports attribution quality across both models and domains.
We compare our attention-based method (\textsc{ClinTrace}) against four baselines:
model self-attribution, where the model itself is prompted to generate its own citations;
embedding-based attributions using the model's own text and image encoders and CLIP; and a random baseline.

\subsubsection{Text-only Attribution}

On text-only samples (1{,}405 from \textsc{CliConSummation} and 1{,}603 from \textsc{MIMIC-CXR}), \textsc{ClinTrace} substantially outperforms all baselines across both backbones.

For \textbf{Qwen3-8B}, our method achieves \textbf{92.15} Text~F1 and \textbf{75.64} Exact Match on radiology reports, and \textbf{88.48} / \textbf{60.60} on dialogue summarization.
For \textbf{HuatuoGPT-Vision-7B}, the improvements are even more pronounced: \textbf{97.22} / \textbf{92.24} on radiology and \textbf{94.09} / \textbf{58.23} on dialogue.
Notably, the medical-finetuned model produces higher attribution quality than the general-purpose model, suggesting that domain-specific training yields more structured and interpretable attention patterns.

All non-attention baselines perform poorly, with self-attribution and embedding methods achieving Text~F1 in the low 30s and EM below 5 in most settings---only marginally above random assignment.
This confirms that neither prompting the model for its own citations nor relying on surface-level embedding similarity provides reliable source grounding.

\subsubsection{Multimodal Attribution}

On the 257 multimodal dialogue samples from \textsc{CliConSummation}, we additionally evaluate image attribution accuracy (Image~F1).
\textsc{ClinTrace} achieves \textbf{82.29} Image~F1 for Qwen3-8B and \textbf{80.53} for HuatuoGPT-Vision-7B, outperforming all baselines.
Embedding-based methods achieve moderate image attribution (54--68 Image~F1) but fail to jointly ground both text and image evidence.

\paragraph{Summary.}
Across both backbones, both domains, and both modalities, attention-based attribution consistently and substantially outperforms all baselines.
The results validate that decoder attentions---when properly aggregated---provide a reliable, training-free mechanism for fine-grained source citation in clinical summarization.

\begin{table*}[ht]
\small
\centering
\caption{
Attribution performance across two clinical summarization domains.
\textsc{ClinTrace} denotes our attention-based attribution method using mean-pooled attention across all layers and heads.
Post-hoc baselines include model self-attribution, embedding-based similarity (processor embeddings and CLIP), and random assignment.
Text~F1 and Text~EM measure macro-averaged F1 and exact match of predicted vs.\ reference source sentence sets.
Image~F1 measures accuracy of correctly citing the image (multimodal dialogue only).
}
\label{tab:attribution}
\begin{tabular}{lccccc}
\toprule
 & \multicolumn{2}{c}{\textbf{Radiology}} & \multicolumn{3}{c}{\textbf{Dialogue}} \\
\cmidrule(lr){2-3} \cmidrule(lr){4-6}
\textbf{Method} & Text F1 & Text EM & Text F1 & Text EM & Image F1 \\
\midrule
\multicolumn{6}{>{\columncolor{gray!15}}l}{\textit{Qwen3-8B}}\\
\textsc{ClinTrace} & \textbf{92.15} & \textbf{75.64} & \textbf{88.48} & \textbf{60.60} & \textbf{82.29} \\
Self-Attribution          & 28.54 & 2.71 & 34.10 & 0.46 & 66.19 \\
Embed$_\text{processor}$  & 34.89 & 4.95 & 39.75 & 0.37 & 58.17 \\
Embed$_\text{CLIP}$       & 31.74 & 3.18 & 37.43 & 0.53 & 54.59 \\
Random                    & 27.34 & 2.80 & 31.04 & 0.44 & 42.99 \\
\midrule
\multicolumn{6}{>{\columncolor{gray!15}}l}{\textit{HuatuoGPT-Vision-7B}}\\
\textsc{ClinTrace} & \textbf{97.22} & \textbf{92.24} & \textbf{94.09} & \textbf{58.23} & \textbf{80.53} \\
Self-Attribution          & 32.63 & 0.54 & 41.93 & 0.92 & 62.10 \\
Embed$_\text{processor}$  & 36.33 & 0.86 & 38.54 & 0.63 & 68.24 \\
Embed$_\text{CLIP}$       & 33.73 & 0.57 & 33.54 & 0.28 & 59.40 \\
Random                    & 27.97 & 2.25 & 30.31 & 0.55 & 48.20 \\
\bottomrule
\end{tabular}
\end{table*}

\subsection{Hallucination Detection via Groundedness}
\label{sec:results_hallucination}

Having established that attention-based attribution is aligned, we now ask: can the \emph{same} attention signals detect hallucinations?
We derive a per-sentence \emph{groundedness score} from the token-level attention weights already computed during attribution (see Section~\ref{sec:method}) and evaluate whether this score discriminates between faithful and unfaithful generated content.

\paragraph{Faithfulness labeling.}
We use an LLM judgeL Claude-Opus-4.6~\citep{anthropic2025claude} to label each generated summary chunk as \emph{supported}, \emph{partially supported}, or \emph{unsupported} with respect to the source document.
For binary evaluation, we treat \emph{supported} as faithful and \emph{partially supported} + \emph{unsupported} as unfaithful.
Both models exhibit similar base rates of unfaithful content: Qwen3-8B produces 37.8\% unfaithful chunks (1{,}272 / 3{,}369), while HuatuoGPT-Vision-7B produces 38.3\% (1{,}150 / 3{,}000).

\subsubsection{Groundedness Predicts Faithfulness}
\label{sec:results_auroc}

Table~\ref{tab:hallucination} reports AUROC and Precision@20 (P@20) for hallucination detection.
We compare groundedness scores against a strong embedding confidence baseline (maximum cosine similarity between each generated chunk and any source chunk using a state-of-the-art model \texttt{all-MiniLM-L6-v2} trained from classic Sentence-BERT~\citep{reimers2019sentence} architecture) and a random baseline.

\begin{table}[t]
    \small
    \centering
    \caption{Hallucination detection performance. AUROC and Precision@20 for distinguishing faithful from unfaithful generated chunks. 
    For groundedness, we select the best-performing aggregation per model (Max for Qwen3-8B; Mean for HuatuoGPT-7B; see Table~\ref{tab:groundedness_ablation} for the full ablation).}
    \label{tab:hallucination}
    \begin{tabular}{lcccc}
    \toprule
    & \multicolumn{2}{c}{\textbf{Qwen3-8B}} & \multicolumn{2}{c}{\textbf{HuatuoGPT-7B}} \\
    \cmidrule(lr){2-3} \cmidrule(lr){4-5}
    \textbf{Method} & AUROC $\uparrow$ & P@20 $\uparrow$ & AUROC $\uparrow$ & P@20 $\uparrow$ \\
    \midrule
    Groundedness (Ours) & 0.586 & \textbf{0.750} & \textbf{0.767} & \textbf{1.000} \\
    Embedding confidence & \textbf{0.697} & 0.500 & 0.764 & \textbf{1.000} \\
    Random & 0.504 & 0.400 & 0.507 & 0.450 \\
    \bottomrule
    \end{tabular}
\end{table}

For \textbf{HuatuoGPT-Vision-7B}, groundedness achieves \textbf{0.767 AUROC}, slightly outperforming embedding confidence (0.764) while requiring \emph{zero additional computation} beyond what the attribution pipeline already produces.
P@20 reaches 1.0 for both methods, indicating that the 20 least-grounded chunks are all genuinely unfaithful.

For \textbf{Qwen3-8B}, groundedness is weaker (0.586 AUROC), underperforming embedding confidence (0.697).
We attribute this gap to a fundamental difference in how general-purpose and medical-finetuned models distribute attention: HuatuoGPT, trained on medical visual data, develops sharper, more semantically structured attention over clinical content, making the concentration of attention a reliable proxy for factual grounding.
Qwen3-8B, as a general-purpose model, may distribute attention more diffusely even over well-grounded content, reducing the discriminative power of the groundedness signal.

Despite this difference, Qwen3-8B's groundedness score still achieves \textbf{P@20 = 0.75}, indicating three out of four flagged chunks are truly unfaithful, which remains clinically useful for targeted review.

\paragraph{Aggregation ablation.}
Table~\ref{tab:groundedness_ablation} reports AUROC across six aggregation strategies for computing sentence-level groundedness from token-level scores.

\begin{table}[t]
    \centering
    \small
    \caption{Effect of groundedness score aggregation on hallucination detection (AUROC). Best aggregation per model is bolded.}
    \label{tab:groundedness_ablation}
    \begin{tabular}{lcccc}
    \toprule
    & \multicolumn{2}{c}{\textbf{Qwen3-8B}} & \multicolumn{2}{c}{\textbf{HuatuoGPT-7B}} \\
    \cmidrule(lr){2-3} \cmidrule(lr){4-5}
    \textbf{Aggregation} & AUROC & P@20 & AUROC & P@20 \\
    \midrule
    Mean & 0.518 & 0.650 & \textbf{0.767} & \textbf{1.000} \\
    Max & \textbf{0.586} & \textbf{0.750} & 0.727 & \textbf{1.000} \\
    Median & 0.476 & 0.650 & 0.763 & \textbf{1.000} \\
    Min & 0.451 & 0.350 & 0.752 & \textbf{1.000} \\
    Std & 0.390 & 0.050 & 0.275 & 0.050 \\
    Mean$-$Std & 0.423 & 0.350 & 0.748 & 0.750 \\
    \bottomrule
    \end{tabular}
\end{table}

For HuatuoGPT-7B, all location-based aggregations (Mean, Max, Median, Min) perform similarly well (AUROC 0.73--0.77), with Mean slightly leading.
This suggests that the medical-finetuned model's attention is uniformly informative across tokens within a sentence.
For Qwen3-8B, Max is the only aggregation that meaningfully exceeds random, indicating that the peak attention signal---rather than the average---carries the hallucination-relevant information in general-purpose models.
Dispersion-based measures (Std, Mean$-$Std) perform poorly for both models, confirming that \emph{attention concentration}, not attention consistency, is the operative signal.

\subsubsection{Connection to Attribution Quality}
\label{sec:results_connection}

\begin{figure}[t]
    \centering
    \includegraphics[width=1\linewidth]{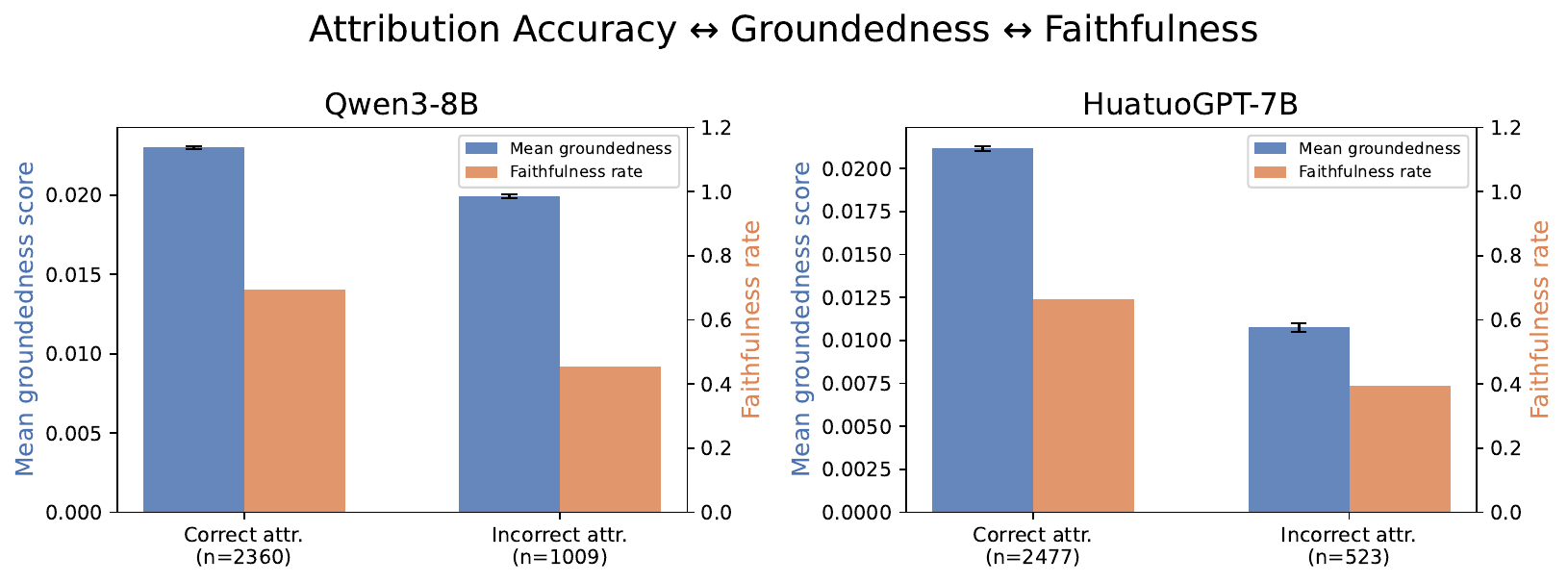}
    \vspace{-20pt}
    \caption{\textbf{Attribution accuracy $\leftrightarrow$ groundedness $\leftrightarrow$ faithfulness.}
    Correctly attributed chunks exhibit both higher mean groundedness scores and higher faithfulness rates than incorrectly attributed chunks, across both models.
    Error bars indicate 95\% confidence intervals.
    This validates the core insight: attribution, groundedness, and faithfulness are three views of the same underlying signal.}
    \label{fig:connection}
\end{figure}

Figure~\ref{fig:connection} examines whether attribution accuracy, groundedness, and faithfulness are connected---the central claim of our framework.
We partition all generated chunks into those with \emph{correct} attribution (predicted source set matches the reference) and those with \emph{incorrect} attribution, then compare their mean groundedness scores and faithfulness rates.

Across both models, correctly attributed chunks have systematically \emph{higher groundedness} and \emph{higher faithfulness} than incorrectly attributed chunks.
For HuatuoGPT-7B, the separation is particularly clean: correctly attributed chunks ($n{=}2{,}477$) have a mean groundedness of 0.021 and faithfulness rate of 63\%, compared to 0.006 and 42\% for incorrectly attributed chunks ($n{=}523$).
For Qwen3-8B, the pattern holds but with narrower margins (groundedness: 0.023 vs.\ 0.020; faithfulness: 58\% vs.\ 42\%), consistent with the weaker AUROC observed above.

This result validates the central thesis of our framework: \emph{attribution, groundedness, and faithfulness are three views of the same underlying attention signal.}
When the model attends strongly and consistently to specific source evidence (high groundedness), it produces correct attributions and faithful content.
When attention is diffuse (low groundedness), attribution fails and hallucination risk increases.

\subsection{Clinical Utility: Abstention}
\label{sec:results_abstention}

The practical value of hallucination detection depends on whether it can improve the \emph{reliability} of model output in a clinical workflow.
We operationalize this through \emph{abstention}: withholding generated chunks whose groundedness scores fall below a threshold $\phi_{\min}$, and presenting only the retained, higher-confidence output to clinicians.

\subsubsection{Faithfulness--Coverage Trade-off}

\begin{figure}[t]
    \centering
    \includegraphics[width=0.6\linewidth]{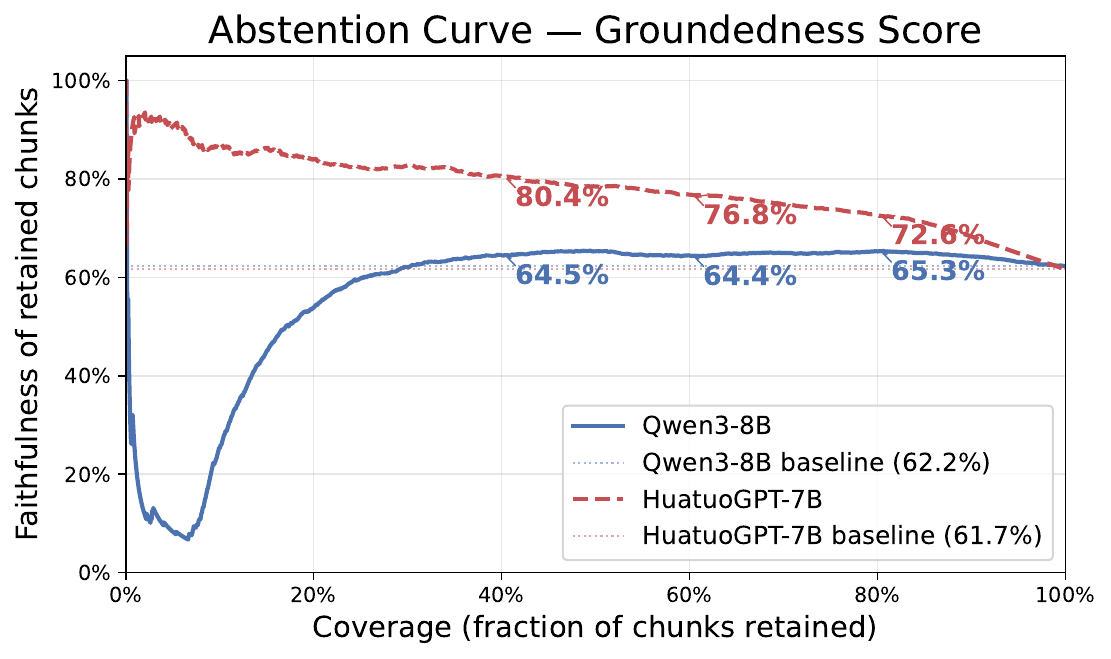}
    \vspace{-5pt}
    \caption{\textbf{Abstention curve.}
    Faithfulness of retained chunks vs.\ coverage as the groundedness threshold $\phi_{\min}$ increases.
    Dotted lines indicate each model's baseline faithfulness at full coverage (no abstention).
    HuatuoGPT-7B shows a smooth, monotonic trade-off: withholding 20\% of output raises faithfulness from 61.7\% to 72.6\%.
    Qwen3-8B exhibits a non-monotonic curve, reflecting the weaker groundedness--faithfulness correlation in general-purpose models.}
    \label{fig:abstention}
\end{figure}

Figure~\ref{fig:abstention} plots the faithfulness of retained chunks against coverage (the fraction of chunks retained) as $\phi_{\min}$ is swept from low to high.

For \textbf{HuatuoGPT-Vision-7B}, the abstention curve is smooth and monotonically increasing as coverage decreases:
at 80\% coverage (withholding the 20\% least-grounded chunks), faithfulness rises from a baseline of 61.7\% to \textbf{72.6\%}---an 11-point improvement.
At 60\% coverage, faithfulness reaches \textbf{76.8\%}, and at 40\% coverage, \textbf{80.4\%}.
This curve provides a practical operating guide: a clinical deployment can choose the desired trade-off between output completeness and reliability.

For \textbf{Qwen3-8B}, the abstention curve is non-monotonic at low coverage, with faithfulness dipping below baseline when fewer than 20\% of chunks are retained.
At moderate coverage ($\geq$40\%), faithfulness stabilizes at 64--65\%, a modest improvement over the 62.2\% baseline.
This behavior is consistent with the weaker AUROC observed in Section~\ref{sec:results_auroc}: the general-purpose model's attention patterns do not provide a reliable ranking of faithfulness at the extremes.

\paragraph{Clinical interpretation.}
These results suggest that abstention is most effective when combined with medical-finetuned models.
For HuatuoGPT-7B, a clinician reviewing a model-generated summary could trust that chunks \emph{not} flagged for review are substantially more reliable than the raw output.
A practical deployment might present retained chunks normally while highlighting abstained chunks with a visual indicator (e.g., a colored background or warning icon), enabling clinicians to allocate their verification effort efficiently. Qualitative examples are in Appendix~\ref{app:qualitative}.
\section{Discussion}
\label{sec:discussion}

\paragraph{Clinical implications.}
Our results suggest that attention-based attribution can serve as a practical audit layer for clinical summarization, with an important caveat: its utility depends substantially on the underlying model.
The most actionable finding for clinicians is that \emph{medical-finetuned models provide substantially more reliable self-auditing signals than general-purpose models}.
With HuatuoGPT-Vision-7B, withholding the 20\% least-grounded output raises faithfulness from 61.7\% to 72.6\%---a meaningful improvement for a zero-cost intervention.
At over 90\% text F1, the attention-based citations are additionally reliable enough to serve as navigational aids, helping clinicians locate source material behind each claim without re-reading the entire input.

\paragraph{Medical finetuning structures attention for interpretability.}
The gap in hallucination detection between HuatuoGPT (0.77 AUROC) and Qwen3 (0.59 AUROC)---models sharing the same architecture family---suggests that medical finetuning reshapes attention distributions over clinical content in ways that improve self-auditing.
This observation has implications beyond our framework: domain adaptation may provide interpretability benefits not captured by standard task performance metrics.

\paragraph{Connection to concurrent work.}
Our attention-based attribution pipeline shares methodological foundations with OmniTrace~\citep{yan2026omnitraceunifiedframeworkgenerationtime}, a concurrent framework for generation-time attribution across omni-modal inputs.
While OmniTrace addresses general omni-modal attribution, our work focuses on clinical deployment and contributes groundedness scoring for hallucination detection and the finding that medical finetuning improves attention-based self-auditing.

\paragraph{Limitations.}
Both attribution reference labels and faithfulness annotations are constructed using an LLM judge (Claude Opus 4.6~\citep{anthropic2025claude}); while validated on a subset through manual inspection, LLM-based evaluation may introduce biases for clinical content involving nuanced reasoning or negation.
Attention-based attribution is correlational, not causal: high attention to a source span is consistent with but does not prove factual grounding.
Our evaluation covers two clinical domains and two model backbones; generalization to other specialties, languages, and model families remains to be validated.
Finally, real-world deployment would require prospective evaluation with clinicians to assess whether these mechanisms improve trust and patient safety---we present evidence for feasibility, not readiness.

\bibliographystyle{unsrtnat}
\bibliography{main}

\newpage
\appendix
\section{Prompts}
\label{appendix:prompts}

\subsection{Task-Specific Generation Prompts}
\label{app:task_prompt}

We use minimal, task-appropriate prompts for each clinical summarization domain. No few-shot examples are provided; all generation uses greedy decoding.

\paragraph{Radiology report summarization (MIMIC-CXR).}

\begin{lstlisting}
Summarize the FINDINGS below into a concise IMPRESSION section:
{findings_text}
\end{lstlisting}

\paragraph{Dialogue summarization (CliConSummation).}

\begin{lstlisting}
Summarize the doctor-patient dialogue below:
{dialogue_text}
\end{lstlisting}

For multimodal dialogues containing an embedded image, the image is provided as a visual input to the model alongside the text prompt, following the model's native multimodal input format.

\subsection{LLM Judge Prompts}
\label{app:llm_judge_prompt}

We use Claude Opus 4.6~\citep{anthropic2025claude} as the LLM judge for two evaluation tasks: attribution reference label construction and faithfulness annotation. The judge model is distinct from both generation models to reduce evaluation bias.

\paragraph{Attribution judge.}
For each generated summary sentence, the judge identifies which source chunks (text and/or image) support it. The system prompt is:

\begin{lstlisting}
You are an expert attribution judge for clinical NLP.
Your task is to determine which input chunks a
generated output sentence is derived from.

You will be given:
- A list of numbered input TEXT chunks (each with a
  chunk_id and text)
- For multimodal samples: one or more input images
  (each labelled with its chunk_id) shown inline
- A single generated output sentence

Return a JSON object with exactly two keys:
- "text_sources_gt": a list of chunk_ids (integers)
  from the text chunks that the sentence draws
  content or meaning from
- "image_sources_gt": a list of chunk_ids (integers)
  from the image chunks that the sentence draws
  visual content from, empty if image is not provided

Rules:
- Include a text chunk if the generated sentence
  directly references, paraphrases, or depends on
  information in that chunk.
- Include an image chunk if the generated sentence
  describes or depends on visual information visible
  in that image.
- The lists may be empty if nothing was drawn from
  that modality.
- Return only valid JSON, no extra text.
\end{lstlisting}

The user message provides the numbered source chunks, any images, and the generated sentence to be judged.

\paragraph{Faithfulness judge.}
For each generated summary sentence, the judge classifies its faithfulness with respect to the full source document. The system prompt is:

\begin{lstlisting}
You are evaluating the faithfulness of a generated
clinical summary sentence.

You will be given a source document and a generated
sentence. Judge whether the sentence is faithfully
supported by the source.

Respond with exactly one word:
- supported
    all claims in the sentence are directly supported
    by the source
- partially_supported
    some claims are supported but the sentence
    contains information not in the source or minor
    inaccuracies
- unsupported
    the sentence contains claims that cannot be
    verified from the source or are fabricated
\end{lstlisting}

The user message provides the full source document followed by the generated sentence to be judged.
\section{Qualitative Examples}
\label{app:qualitative}

We present four examples illustrating how groundedness scores correlate with faithfulness across both clinical domains.
Each example shows the generated summary sentences, their mean groundedness scores ($\phi$), attributed sources, and faithfulness labels from the LLM judge.
All examples use \texttt{Qwen3-8B}.

\subsection{Multimodal Dialogue: Acanthosis Nigricans}

\paragraph{Source (abbreviated).}
A patient reports weight gain. The doctor asks about infertility (confirmed), examines a skin image, asks about warts (confirmed), and diagnoses acanthosis nigricans (Group~12).

\begin{table}[h]
\small
\centering
\begin{tabular}{clcl}
\toprule
\textbf{Sent} & \textbf{Generated text (abbreviated)} & $\phi$ & \textbf{Label} \\
\midrule
0 & ``The patient reports weight gain and infertility.'' & 0.027 & \textcolor{green!50!black}{supported} \\
1 & ``Doctor notes a melanocytic naevus and warts\ldots'' & 0.023 & \textcolor{orange!80!black}{partial} \\
2 & ``Acanthosis nigricans, associated with insulin resistance\ldots'' & 0.020 & \textcolor{orange!80!black}{partial} \\
3 & ``Often linked to PCOS, which may explain infertility\ldots'' & 0.016 & \textcolor{red!70!black}{unsupported} \\
\bottomrule
\end{tabular}
\end{table}

\paragraph{Analysis.}
Sentence~0 is well-grounded ($\phi{=}0.027$) and directly paraphrases the patient's statements~[1] and~[4].
Sentence~1 introduces ``melanocytic naevus''---a clinical term not present in the dialogue---likely inferred from the image but not stated by the doctor, producing a lower score.
Sentence~2 adds ``insulin resistance and metabolic syndrome,'' medical knowledge not mentioned by either party.
Sentence~3 is entirely fabricated: the dialogue never mentions PCOS, and the model generates a plausible but unsupported clinical explanation.
The monotonic decrease in $\phi$ (0.027~$\to$~0.016) tracks the progression from faithful summarization to hallucinated medical reasoning.

\paragraph{Clinical significance.}
A clinician reading sentence~3 might accept the PCOS explanation at face value, potentially influencing workup decisions (e.g., ordering hormone panels) based on a fabricated association.
The low groundedness score correctly flags this sentence for review.

\subsection{Multimodal Dialogue: Diabetes Insipidus}

\paragraph{Source (abbreviated).}
A patient presents with cough and concern about upcoming exams. The doctor asks about excessive appetite (confirmed), difficulty swallowing (confirmed), frequent urination (confirmed), and itchy eyelid (uncertain---patient shares an eye image). The doctor diagnoses diabetes insipidus (Group~4).

\begin{table}[h]
\small
\centering
\begin{tabular}{clcl}
\toprule
\textbf{Sent} & \textbf{Generated text (abbreviated)} & $\phi$ & \textbf{Label} \\
\midrule
0 & ``Patient presents with cough, concern about exams.'' & 0.025 & \textcolor{green!50!black}{supported} \\
1 & ``Excessive appetite, difficulty swallowing, urination confirmed.'' & 0.023 & \textcolor{orange!80!black}{partial} \\
2 & ``Doctor observes eye via image $\to$ diabetes insipidus.'' & 0.018 & \textcolor{orange!80!black}{partial} \\
3 & ``Cough is likely a secondary symptom of underlying condition.'' & 0.015 & \textcolor{red!70!black}{unsupported} \\
\bottomrule
\end{tabular}
\end{table}

\paragraph{Analysis.}
Sentence~3 is the clearest hallucination: the doctor never characterizes the cough as secondary, and no causal relationship between cough and diabetes insipidus is stated in the dialogue.
The model generates a clinically plausible but entirely unsupported inference.
Again, $\phi$ decreases monotonically from the well-grounded opening sentence to the fabricated conclusion.

\paragraph{Clinical significance.}
Presenting the cough as ``secondary'' could lead a clinician to deprioritize further investigation of the cough, potentially missing an independent respiratory condition.

\subsection{Radiology: Port-A-Cath / Chest X-ray}

\paragraph{Source.}
\emph{``Right-sided Port-A-Cath terminates in the mid SVC as before. Heart is top-normal in size. Mediastinal and hilar contours are within normal limits. Lung volumes are low over the lungs are clear without focal consolidation, effusion or pneumothorax.''}

\begin{table}[h]
\small
\centering
\begin{tabular}{clcl}
\toprule
\textbf{Sent} & \textbf{Generated text (abbreviated)} & $\phi$ & \textbf{Label} \\
\midrule
0 & ``IMPRESSION:'' (header only) & 0.034 & \textcolor{red!70!black}{unsupported} \\
1 & ``Port-A-Cath appropriately positioned in mid-SVC.'' & 0.035 & \textcolor{green!50!black}{supported} \\
2 & ``Normal heart size and mediastinal/hilar contours.'' & 0.023 & \textcolor{green!50!black}{supported} \\
3 & ``Bilateral lung fields clear with reduced volumes\ldots'' & 0.021 & \textcolor{orange!80!black}{partial} \\
\bottomrule
\end{tabular}
\end{table}

\paragraph{Analysis.}
Sentence~0 (``IMPRESSION:'') is an interesting edge case: the model outputs a section header that is technically unsupported content, but its high $\phi$ reflects strong attention to the prompt instruction rather than clinical evidence.
This represents a limitation of the LLM judge---trivial formatting tokens may be labeled as unsupported even though they are clinically harmless.
Sentence~3 introduces ``bilateral,'' but the source only states ``lungs are clear'' without laterality---a subtle inaccuracy that could matter if asymmetric findings were present in the original images.

\subsection{Radiology: PICC Line / Pleural Effusion}

\paragraph{Source.}
\emph{``Right PICC line ends at low SVC. Moderate right pleural effusion with adjacent lung atelectasis has decreased since \_\_\_. Minimal left pleural effusion is unchanged. There are no new lung opacities of concern for pneumonia. Heart size, mediastinal and hilar contours are stable.''}

\begin{table}[h]
\small
\centering
\begin{tabular}{clcl}
\toprule
\textbf{Sent} & \textbf{Generated text (abbreviated)} & $\phi$ & \textbf{Label} \\
\midrule
0 & ``IMPRESSION:'' (header only) & 0.035 & \textcolor{red!70!black}{unsupported} \\
1 & ``Right PICC line terminates in low SVC.'' & 0.033 & \textcolor{green!50!black}{supported} \\
2 & ``Pleural effusions stable or decreased\ldots'' & 0.022 & \textcolor{orange!80!black}{partial} \\
3 & ``Cardiac and mediastinal structures unchanged.'' & 0.022 & \textcolor{green!50!black}{supported} \\
\bottomrule
\end{tabular}
\end{table}

\paragraph{Analysis.}
Sentence~2 conflates two distinct findings: the right effusion has \emph{decreased} while the left is \emph{unchanged}, but the generated summary merges them as ``stable or decreased.''
This is a clinically meaningful distinction---a decreasing right effusion and a stable left effusion carry different implications for treatment monitoring.
The moderate groundedness score ($\phi{=}0.022$) reflects the model's attention being split across multiple source sentences~[2] and~[3], consistent with the merged and partially inaccurate output.

\paragraph{Summary of patterns.}
Across all four examples, two patterns emerge:
(1)~groundedness scores decrease monotonically or near-monotonically from well-supported opening sentences to less-grounded later sentences, consistent with the known attention dilution phenomenon in autoregressive generation;
(2)~the lowest-groundedness sentences are consistently those containing fabricated clinical reasoning (PCOS association, cough as secondary symptom) or subtle factual conflations (bilateral, stable-vs-decreased).
These are precisely the failure modes most dangerous in clinical deployment, and the ones that groundedness-based abstention is designed to catch.
\section{Data Release}

We will publicly release a comprehensive code base that includes the ClinTrace implementation with different scoring functions. 

We would also release the test set. The licensing terms for the artifacts will follow those set by the respective dataset creators, as referenced in this work, while the curated artifacts will be provided under the MIT License. 
Additionally, our release will include standardized evaluation protocols, and evaluation scripts to facilitate rigorous assessment. The entire project will be open-sourced, ensuring free access for research and academic purposes.

\end{document}